\documentclass[sigconf]{aamas} 

\usepackage{enumitem}
\usepackage{balance} % for balancing columns on the final page
\usepackage{balance}
\usepackage{xcolor}

 \usepackage{tcolorbox}
\usepackage{lipsum}
\usepackage{url}

\usepackage{breakurl}
\usepackage{hyperref}

\usepackage{hyperref}

\setcopyright{ifaamas}
\acmConference[AAMAS '21]{Proc.\@ of the 20th International Conference on Autonomous Agents and Multiagent Systems (AAMAS 2021)}{May 3--7, 2021}{Online}{U.~Endriss, A.~Now\'{e}, F.~Dignum, A.~Lomuscio (eds.)}
\copyrightyear{2021}
\acmYear{2021}
\acmDOI{}
\acmPrice{}
\acmISBN{}

\acmSubmissionID{Blue Sky \#9}

\title{Diverse Auto-Curriculum  is Critical for Successful \\ Real-World Multiagent Learning Systems}
\titlenote{The authors thank the anonymous reviewers, as well as Greg d'Eon, Calarina Muslimani, Laura Petrich, Sahir, and Amirmohsen Sattarifard for comments and suggestions.
Part of this work has taken place in the Intelligent Robot Learning (IRL) Lab, which is supported in parts by research grants from Amii, CIFAR, and NSERC.}
\subtitle{Blue Sky Ideas Track}

\author{Yaodong Yang}
\affiliation{
  \institution{University College London}
    \institution{Huawei R\&D U.K.}
}
\authornote{Corresponding author: \texttt{yaodong.yang@huawei.com}}

\author{Jun Luo}
\affiliation{
  \institution{Huawei Canada}
}
\author{Ying Wen}
\affiliation{
  \department{Shanghai Jiao Tong University}
}
\author{Oliver Slumbers}
\affiliation{
  \department{University College London}
}
\author{Daniel Graves}
\affiliation{
  \department{Huawei Canada}
}
\author{Haitham Bou Ammar}
\affiliation{
  \department{Huawei R\&D U.K.}
}
\author{Jun Wang}
\affiliation{
  \department{University College London}
  \institution{Huawei R\&D U.K.}
}
\author{Matthew E. Taylor}
\affiliation{
  \department{University of Alberta} %\& Amii}
  \department{Alberta Machine Intelligence Institute}
}
\begin{abstract}

Multiagent reinforcement learning (MARL) has achieved a remarkable amount of success in solving various types of video games. A cornerstone of this success is the auto-curriculum framework, which shapes the learning process by continually creating new challenging tasks for agents to adapt to, thereby facilitating the acquisition of new skills. In order to extend MARL methods to  real-world domains outside of video games, we envision in this blue sky paper that maintaining a diversity-aware auto-curriculum is critical for successful MARL applications. Specifically, we argue that \emph{behavioural diversity} is a pivotal, yet under-explored,  component for real-world multiagent learning systems, and that significant work remains in understanding how to design a diversity-aware auto-curriculum. We list four open challenges for auto-curriculum techniques,  which we believe deserve more attention from this community. Towards validating our vision, we recommend modelling realistic interactive behaviours in  autonomous driving as an important test bed, and recommend the SMARTS/ULTRA benchmark.

\end{abstract}


\newcommand{\BibTeX}{\rm B\kern-.05em{\sc i\kern-.025em b}\kern-.08em\TeX}

\begin{document}

%%% The following commands remove the headers in your paper. For final 
%%% papers, these will be inserted during the pagination process.

\pagestyle{fancy}
\fancyhead{}

%%% The next command prints the information defined in the preamble.

\maketitle 

%%%%%%%%%%%%%%%%%%%%%%%%%%%%%%%%%%%%%%%%%%%%%%%%%%%%%%%%%%%%%%%%%%%%%%%%

%\blfootnote{$^1$University College London, $^2$Huawei R\&D U.K., $^3$Huawei Canada, $^4$Shanghai Jiao Tong University, $^5$University of Alberta, $^6$Alberta Machine Intelligence Institute (Amii)}

%\vspace{-7pt}
\section{Introduction}
\label{sec:Intro}
Reinforcement learning (RL) \cite{sutton2018reinforcement} allows an agent to learn to maximise cumulative rewards via environmental interactions. It has proved successful in many areas, including playing video games \cite{mnih2013playing,peng2017multiagent}, robotics control \cite{lee2020learning}, data centre cooling \cite{li2019transforming}, and asset pricing in finance \cite{kolm2020modern}. Multiagent RL (MARL) extends RL to cover the setting where there are multiple learning entities in the environment \citep{yang2020overview, hernandez2019survey}. This technique has shown remarkable success,   especially on multi-player video games such as StarCraft \cite{vinyals2019alphastargrandmaster}, Dota2 \citep{pachocki2018openai}, and Hide and Seek \cite{baker2019emergent}. However, MARL has had relatively few successes in solving real-world problems. The core thesis of this paper is that 
%in order to help MARL succeed in real-world domains, developing 
the development of learning frameworks that can induce  \emph{behavioural diversity} in the policy space is critical for MARL to succeed in real world domains.
%though this issue is as yet under-explored by the MARL community. 
We summarise existing challenges and recommend  autonomous driving (AD) as an ideal test bed for future investigations.

 %\MET{Note - if genetic algorithm people review this, they may jump on us. Weaken here/specify why GAs or other population-based methods are a great start, but are not sufficient?}

The challenges of deploying RL in the real world are frequently discussed in both workshops \cite{nips-workshop,rl4life2,rl4life} and papers \cite{dulac2020empirical,dulac2019challenges}. Challenges include the lack of an accurate simulator \cite{boedecker2008simspark}, the high cost of environmental interaction, and the difficulty in learning both effective and diverse policies. 
Off-policy RL \cite{levine2020offline} or imitation learning \citep{hussein2017imitation}  methods could be used for policy evaluation or policy improvement in an offline manner; this would allow agents to learn a good initial behaviour before ever interacting with an environment. However, these methods are only applicable if training data sets exist. %\MET{return to this later - a non-diverse data set won't be useful}
When the training data  are limited, for example in the AD domain  \citep{zhan2019interaction}, offline methods are insufficient for robust performance in the real world  due to a lack of \emph{diversity} in agents' behaviours \cite{dasari2019robonet,wang2017robust}. 
In fact, even in cases when a simulator is available, lack of behavioural diversity could still exist due to the sim-to-real gap    \cite{rusu2017sim,openai2018dactyl,matas2018sim}. 
%Another common approach is to  train models in a simulator, if available \ollie{I think we need to be more careful with the wording here, as we've stated lack of simulators to be a challenge, and yet stated the fix of the challenege is just that simulators are a common approach}. However,   
%\ollie{I may be confused, but I am not sure what sim-to-real has to do with the diversity issue as this sentence seems to be mentioning}. 
%The robotics and RL communities are well aware of this issue and try to address it by leveraging sim2real methods , or robust  algorithms \citep{morimoto2005robust,pinto2017robust,abdullah2019wasserstein}. 
%We know that all models are wrong, but some are useful---the robotics and RL communities are well aware that pre-training on an imperfect simulator can still cause significant improvements in real-world performance, possibly by 
Unfortunately, MARL suffers from all of these concerns. 
%And things only get worse when one considers 
Furthermore, the additional complexity of multi-agent problems that  arises from the  cross product  of multiple  agents' state and action spaces induced by social interactions compounds these concerns. 
Developing frameworks that can deal with the underlying complexities of the MARL domain is crucial. We argue that the development of effective, yet diverse behaviours is critical for MARL to have an impact in domains outside of video games.

The \emph{auto-curricula framework} \citep{leibo2019autocurricula,portelas2020automatic} is a promising direction towards such a goal.
%Towards such a goal, the framework of \emph{auto-curricula} \citep{leibo2019autocurricula,portelas2020automatic} serves as a promising direction. 
In natural evolution, species with stronger adaptability flourish when nature alters the environment and some previously well-adapted species no longer survive. 
%\MET{Should we mention that a library of past policies can be kept around to ensure behaviours don't cycle, but instead improve over time? This would also help prevent catastrophic forgetting.}
Through this \emph{co-evolution process} \cite{rausher2001co,durham1991coevolution,paredis1995coevolutionary}, the diversity of life on Earth has been maintained over billions of years.
Inspired by this mechanism of bio-diversity in nature, a series of MARL learning frameworks have recently been proposed and have demonstrated remarkable empirical success. These include open-ended evolution \citep{standish2003open, banzhaf2016defining,lehman2008exploiting}, population based training  \citep{jaderberg2019human,liu2018emergent}, and training by emergent curricula \citep{leibo2019autocurricula, baker2019emergent, portelas2020automatic}. 
%have been proposed  and  demonstrated remarkable empirical success in solving complicated real-world games. 
In general, these frameworks can be unified under the idea of an auto-curriculum where an endless procession of  better-performing agents are automatically generated by exerting selection pressure among the multiple self-optimising agents. 
The underlying principle of auto-curricula is that any adaptation  an agent makes will have a cascading effect that other agents must adapt to in order to survive. This intrinsically provides 
%there exists an intrinsic mechanism, arising from agents' social interactions,
%(either competitions or cooperations), 
that provides an automatic curriculum that continually  facilitates agents' acquisition of new skills via such social interactions. 
%them to acquire new skills that would not be learnable otherwise. 
%The inherent curriculum in multiagent systems  (hereafter  \emph{auto-curriculum}) provides a continual loop in creating challenges that push agents to keep adapting. 
%Such a  self-generated sequence of  challenges from multiagent co-adaptations  is also called \emph{auto-curriculum} \cite{leibo2017multi}.

In this Blue Sky paper, 
we emphasise that maintaining a diversity-aware auto-curriculum is critical for successful MARL applications. Specifically, we advocate verification through one specific real-world problem: modelling interactive behaviours in   AD scenarios.    
The main contributions of this work are as follows.  
We start by highlighting the necessity of behavioural diversity in multiagent systems in Section~\ref{sec:diversity} and  then briefly survey  existing works  and explain why they are not yet sufficient for MARL  in Section~\ref{sec:related}.
We investigate the idea of auto-curriculum in Section~\ref{sec:solutions}, and raise four  open challenges which we believe deserve more attention from this community. 
Section~\ref{sec:SMARTS} discusses why AD is an excellent domain to host such investigations and proposes one particular test bed for future study.
Finally, we reiterate our vision in Section~\ref{sec:conclusion}. %next steps for our team and the recommunity to tackle this important and challenging problem. %\haitham{Really like this but a summary of what MARL problems are beyond diversity and saying we choose to focus on diversity is important I think. To me, and it can only be me, this sounds like solve diversity you get real-world MAS but we know that's not the case.}
\vspace{-2pt}
\section{The Necessity of Diversity}
\label{sec:diversity}
%\MET{Define and highlight the need for diversity (zero-sum games, Hanabi, and driving) (Section 2). Yaodong leads, 0.5 page.}
%Matt talks about Hanabi} \yaodong{I update the paragraph, maybe too long, please help shrink if needed}
%\MET{We may need to cut/trim the biology references if we need the space.}
Nature exhibits a remarkable tendency towards \textit{diversity}  \citep{holland1992adaptation}. 
Over the past billions of years, a vast assortment of unique species have naturally evolved. Each one is capable of orchestrating the complex biological processes necessary to sustain life. 
Analogously, in computer science, machine intelligence can be considered as the ability to adapt to a diverse set of complex environments \citep{hernandez2017measure}. 
This suggests that the ceiling of intelligence rises when environments of increasing diversity and complexity are provided.
In fact, recent successes in developing AI capable of achieving super-human performance on complicated multi-player video games, such as StarCraft  
\cite{vinyals2019grandmaster,han2020tstarbot},  Honour of King \cite{ye2020towards}, Hide and Seek \citep{baker2019emergent}, and Dota2 \cite{pachocki2018openai}, have  provided justification for emphasising behavioural diversity  when designing learning protocols in  multiagent systems. 
%\MET{Yaodong: I'm not seeing why these show diversity is important --- do these papers rely on it?}\yaodong{the angel in starcraft is that, they have to find a very diverse set of behaviors, through different types of exploiters which keep exploting the existing strategy pool to push for diversity,  so as not to be easily exploited when playing against humans} 
Specifically, promoting behavioural diversity is pivotal for MARL methods. Diversity not only prevents AI agents from checking the same policies repeatedly, but also helps agents discover niche skills, avoid systematic weaknesses and maintain robust performance when encountering unfamiliar types of  opponents at test time.    
Behavioural diversity and the \emph{non-transitivity} of many environments are intertwined. In biological systems, bio-diversity is promoted by  the non-transitive interactions among many competing populations \citep{kerr2002local, reichenbach2007mobility}. The central feature of such non-transitive relations can be thought of as analogous to a Rock-Paper-Scissors game, where rock beats scissors, scissors beats paper, and paper beats rock.
%The natural law indicates that
%\MET{Yaodong: Have we defined diversity and non-transitivity in the context of MARL?} \yaodong{this is a good question. behavorial diversity in RL so far is mostly define on the reward or the visitation of states, if the performance/or the visited state is very different, then it is defined as diverse. I add some jsutification in section 3 that there is no uniform definition of diveristy.  non-transitivity is defined in games yes. which is in the following sentence}
In game theory, the necessity of pursuing behavioural  diversity is also deeply rooted in the non-transitive structure of games \citep{lanctot2017unified, balduzzi2018re, balduzzi2019open}. 
In general, an arbitrary game, of either the normal-form type \citep{candogan2011flows} or the differential type \citep{balduzzi2018mechanics}, can always be decomposed into a sum of two components: a \emph{transitive part} plus a \emph{non-transitive part}. 
The transitive part of a game represents the structure in which the rule of winning is transitive (i.e., if strategy A beats B, B beats C, then A can surely beat C), and the non-transitive part refers to the game structure in which the set of strategies follow a cyclic rule (e.g., the endless cycles of rock, paper, and scissors). 
Diversity matters especially for the non-transitive part because there is no consistent winner in such sub-games: if a player only plays rock, he can be exploited by paper, but not so if he is diverse in playing pock and scissors.  %\ollie{also I think the term exploited needs to be defined in some manner, I feel like it could be interpreted in a couple ways whereas I think we use it strictly in the manner of Exploitability, i.e. is the best response to said strategy a large unilateral deviation or not} % in the RPS game for example. 
In fact, real-world problems often consist of a mixture of both  parts \citep{czarnecki2020real}, 
therefore it is critical to design learning objectives that lead to behavioural diversity. % is critical for tackling these games. %\jun{not sure about this. I thought it is the randomisation that plays the role???}. 

% where we need diversity
%, a measure of diversity is expected to quantify both different and  effective behaviours. After all, we are  neither interested in differences in policies that  do not lead to differences in outcomes, nor in agents that lose in new and surprising ways against opponents. 

Effective MARL performance often requires diversity in two aspects. The first aspect is about the training player uses diversified strategies against a fixed type of opponents. Most games involve non-transitivity in the policy space and thus it is necessary for each player to acquire a diverse set of winning strategies to achieve high unexploitability. 
The second aspect is the ability to pick a diverse set of opponents\footnote{We use the term ``opponent'' for presentation purposes, acknowledging that agents may be teammates, opponents, or something in between for non-zero sum games.} during training. %, which can be justified by many real-world examples. 
%In many real-world tasks, such as autonomous driving, there are uncountably many possible types of opponent behaviours, yet the computational resources are finite. Picking a diverse subset of opponents during training can effectively and efficiently improve the robustness of agents, particularly when the real-world opponents are unknown during training.
In playing cooperative  card games like 
%Arkham Horror, Spirit Island, or Hanabi, players work together in a non-zero sum setting, with or without communication. In all cases, players may have different strategies and assumptions. For example, in 
Hanabi  \citep{bard2020hanabi}, one player may or may not understand the indirect signalling when choosing a card to play. If an agent has not learned to play diversely under both mindsets, it will fail to accurately model the collaborator and play sub-optimally. 
Similarly, in real-world driving, distinct locales have different  conventions. 
The UK and the US drive on different sides of the road, or even within the same country, different cities can follow different conventions. 
For example, the \emph{Pittsburgh Left} convention assumes that a few cars will turn left in front of traffic at the beginning of a green light \citep{pitts_left}, while other areas assume that cars will turn left in front of traffic during a yellow or at the beginning of a red light \cite{red-light}. %Even within a given locale, different people will likely have different preferences/behaviours, and these could change over time. 
As a result, it can be expected that an autonomous agent without a diverse mindset could easily create hazards on the road \cite{waymo2020safety}. 
%depending on the urgency of their current task.

%\vspace{-5pt}
\section{Related Work on Diversity}
\label{sec:related}
%\MET{Discuss how existing diversity methods fall short (Section 3). Yaodong leads .5/.75 page}
%\yaodong{pls check, feel free to amend directly}
Despite the importance of diversity, there has been limited work within the machine learning domain where diversity is modelled in a principled way. Furthermore, there is no agreed upon, formal definition.
%\MET{Yaodong: Can we add a rigorous definition, either in Section 2 or Section 4?}\yaodong{i add the following setence}
%In fact, an unanimous definition of diversity itself is even lacking. 
%Existing definitions are often ad hoc. 
For example, behavioural diversity can be defined as the variance in rewards \citep{lehman2008exploiting,lehman2011abandoning}, the convex hull of a \emph{gamescape} \citep{balduzzi2019open}, choosing whether or not to visit a new environmental state  \citep{yang2020multi,eysenbach2018diversity, such2017deep},  or, acquiring  new types of skills in a task \citep{florensa2017stochastic,hausman2018learning}.

So far, the majority of work that models diversity lies in  evolutionary computation (EC) \citep{fogel2006evolutionary, back1997handbook}, which attempts to mimic the   natural evolution process. %, specifically, they  focus on diversification during the optimisation process. 
One classic idea in EC is \emph{novelty search}  \citep{lehman2008exploiting,lehman2011abandoning}, which aims to search    for behaviours that lead to different outcomes.  %without any underlying objective pressure.
Quality-diversity (QD) methods hybridise novelty search with fitness under the notion of survival of the fittest (i.e., high utility)  \citep{pugh2016quality}. Two representatives are \emph{Novelty Search with Local Competition} \citep{lehman2011evolving} and  \emph{MAP-Elites} \citep{cully2015robots,mouret2015illuminating}.

Searching for behavioural diversity is also a common topic in  RL, which is  often studied in the context of skill discovery \citep{eysenbach2018diversity,florensa2017stochastic,hausman2018learning}, intrinsic rewards \citep{gregor2017variational,bellemare2016unifying,barto2013intrinsic}, or  maximum-entropy learning \citep{haarnoja2017reinforcement,haarnoja2018soft,levine2018reinforcement}. 
These RL algorithms can be considered as QD methods, in the sense that quality refers to maximising cumulative reward, and diversity means either visiting a new state \citep{yang2020multi,eysenbach2018diversity} or  obtaining a policy with larger entropy \citep{levine2018reinforcement}.
%As a result, they still suffer from the same issues of either being an ad hoc solution or having no theoretical justification. 

In the context of MARL, learning typically means an agent acting in an open-ended system with continually changing policies by different opponents. %, and agents are guided by an adaptive objective to keep improving endlessly. 
Yet, such a learning process can only guarantee \emph{differences} but not \emph{diversity}, 
which are two different notions--diversity is not an inherent feature in MARL.
In fact, understanding the principle of how diversity is promoted in an auto-curriculum is an open problem in MARL  \citep{balduzzi2019open,yang2020multi,parker2020effective}.
In the example of training soccer AIs \cite{kurach2020google}, learning against only \emph{different} opponents can easily make an agent get into circular dynamics and not improve. 
Finally, this work is also different from the previous manifesto \citep{leibo2019autocurricula}, which links the auto-curricula in natural evolution with  MARL;  our main focus is to emphasise creating diversity-aware auto-curricula. 
\section{Existing Open Challenges}
\label{sec:solutions}
Auto-curricula \cite{leibo2019autocurricula, baker2019emergent, portelas2020automatic}  provide a framework to automatically  shape learning procedures for AI agents by consistently challenging them  with new tasks that are adapting to their capabilities. 
As the challenges generated by an auto-curriculum 
%In auto-curriculum,  if the generated challenges  
become increasingly diverse and complex over time, AI agents accumulate more diverse and effective skills.
%This suggests that the ceiling of intelligence can be raised by changing the underlying  environments with increasing diversity and complexity, an idea also called \emph{exploration by exploitation}\footnote{Note that this is distinct from the tradeoff of exploration vs. exploitation  in classic RL.} \cite{leibo2017multi}. 
%When a new form of intelligent model emerges, it shifts the environment dynamics and creates new pressure for existing intelligent models to adapt. 
In fact, recent successes in training AIs that achieve super-human performance and acquire diverse behaviours on complex video games \citep{silver2016mastering,vinyals2019grandmaster,baker2019emergent} provide strong justification for adopting auto-curricula a diversifying  learning protocol.  
However, in order to serve as a general  framework to tackle more real-world problems beyond video games, auto-curriculum technique still faces four open challenges. 
% In particular, 
 %Strong evidence has been found on the growing complexity of emergent behaviours resulting from the auto-curriculum, which qualitatively matches human-level or even super human-level skills. 
%We believe the answer to the above three questions is \textbf{diversity}. 
{{\begin{tcolorbox}[fonttitle=\normalsize,fontupper=\normalsize,fontlower=\normalsize,top=1pt,bottom=1pt,left=1pt,right=1pt,title=Open Challenges of Designing  Diversity-Aware Auto-Curricula]
\begin{enumerate}[leftmargin=15pt]
\item	How do we measure diversity in an auto-curriculum? 
\item How do we generate diversity-aware auto-curricula, especially in non-zero sum settings?
\item How do we shape an auto-curriculum to induce diverse yet effective behaviours?
\item How do we deal with non-transitivity when learning in a diversity-aware auto-curricula?
\end{enumerate}
\end{tcolorbox}}}
The first challenge is to define the correct objective to  measure and promote  diversity in the generated auto-curriculum.  
In the single-agent setting, diversity can be defined through a different reward function \citep{lehman2008exploiting,lehman2011abandoning}, visiting a new state \citep{yang2020multi,eysenbach2018diversity, such2017deep}, or acquiring a new skill \citep{florensa2017stochastic,hausman2018learning}. However, in the MARL setting, with multiple players, each having a population of strategies, diversity should be defined in the joint policy space, considering all existing strategies of all agents. Yet, there is limited work that tries to quantify the behavioural diversity at the population level. 
Although there are no straightforward answers, we believe one promising direction could be to  leverage the \emph{determinantal point process} \citep{kulesza2012determinantal} from quantum physics. This processs measuring diversity
%where the diversity is measured 
through  the determinant value in a vector space,  thus the level of orthogonality among the input vectors can be represented by agents' different joint-strategy profiles in terms of rewards \cite{yang2020multi}.
% the probability of sampling a subset from a ground set is modelled by the determinant value of of a correlation kernel so that only the most diverse subsets are likely to be sampled. \MET{I didn't follow this final sentence, and it's pretty important.}
%{\begin{tcolorbox}[fonttitle=\small,fontupper=\small,top=1pt,bottom=1pt,left=1pt,right=1pt,fontlower=\small,title=Open Challenge II]
%How do we generate diversity-aware auto-curricula, especially in non-zero sum settings?
%\end{tcolorbox}}

The second challenge involves the applicability on \emph{non-zero sum games}. 
The curricula in the examples of StarCraft \cite{vinyals2019alphastargrandmaster} or Hide and Seek \cite{baker2019emergent} are  generated by competitive self-play from the players in zero-sum games.
However, many real-world tasks, such as autonomous driving, are not zero-sum---in fact, they tend to be a mixed setting where cooperation outweighs competition. 
Therefore, creating an auto-curriculum in non-zero sum games is an open problem.   
Interestingly, recent studies have shown that adapting in social dilemmas  can create an effective auto-curriculum for the emergence of collective cooperation~\cite{hughes2018inequity,leibo2017multi,perolat2017multi}. Through sequences of new challenges in addressing social dilemmas, agents eventually learn to achieve a socially-beneficial outcome.  
This resembles tasks such as discovering collective driving strategies that can mitigate congestion. For example, consider the case of solving Braess's paradox \cite{braess1968paradoxon} (i.e., a typical example in modelling road network and traffic flow)  where agents progressively learn to  sanction those who tend to over-exploit the common resources, thus  creating new curriculum.   
Nonetheless, creating curricula for collective cooperation is still under-developed relative to auto-curricula induced by zero-sum games. 
Importantly, as pointed out by \citet{leibo2019autocurricula}, auto-curricula induced by   social dilemmas could be cursed by the ``no-free-lunch" property: once you resolve a social dilemma in one place, another one crops up to take its place, a problem also known as higher-order social dilemmas
\cite{ostrom2000collective, mathew2017second}. 
%In fact, depending on the level of autonomy, the expected outcome of multiagent learning matches in accordance with the six levels proposed by SAE.  
%For example, the bottom level solution is nothing but following specific conservative  safety rules and ignoring the existence of adaptive  opponents.  The higher level solutions incorporate opponent modelling and taking best response to interact proactively (e.g., to cut in if needed), and the top level solution focuses on maximising the social welfare of all road users (e.g., to change my route in order not to create a future congestion).  We refer the detailed description of each level of multiagent learning outcome to   \citet{zhou2020smarts}. 
%{
%\begin{tcolorbox}[fonttitle=\small,fontupper=\small,top=1pt,bottom=1pt,left=1pt,right=1pt,fontlower=\small,title=Open Challenge III]
%How do we shape an auto-curriculum to induce diverse yet effective behaviours?
%\end{tcolorbox}}

Thirdly, although RL techniques offer insight into how a desirable behaviour can be learned in a fixed environment, it is still unclear how complex and useful behaviours can be best developed, while these behaviours are influencing the environment. 
%which again influence the environment, are developed especially those effective ones that can further drive agents to develop useful skills. 
In fact, it is often the case that the more complex the behaviour, the less likely it is generated completely from scratch \citep{leibo2019autocurricula}. An example is that it is highly unlikely a world-champion level policy is quickly generated by a curriculum when learning to act in complex environments. Moreover, this issue is only exacerbated when multiple agents ($N \gg 2$) are involved to explore the joint-strategy space. %, which is exactly the problem of generating realistic and diverse interac for AD. 
Fortunately, initial progress has been made by works on \emph{Policy Space Response Oracle (PSRO)} \citep{lanctot2017unified,balduzzi2019open,muller2019generalized} where different kinds of \emph{rectifiers} have been proposed to shape the auto-curricula so that effective behaviours with high quality  can be emphasised. For example,  PSRO with a \emph{Nash rectifier} \cite{balduzzi2019open} explores only strategies that have positive Nash support so as to preserve the  strategy strength. 
Despite the empirical success in generating diverse yet effective strategies, PSRO methods only work in solving   symmetric zero-sum games, a limitation highlighted in Open Challenge II. 
%{\begin{tcolorbox}[fonttitle=\small,fontupper=\small,top=1pt,bottom=1pt,left=1pt,right=1pt,fontlower=\small,title=Open Challenge IV]
%How do we deal with the non-transitivity when learning in a diversity-aware auto-curricula?
%\end{tcolorbox}}

Lastly, results on game decomposition suggests that a  game  \cite{candogan2011flows,balduzzi2019open}  generally consists of both \emph{transitive} and \emph{non-transitive} structures.  
%In the transitive part, the law of achieving larger utility enjoys transitive property, that is, if strategy A is better than B, B is better than C, then A is surely better than C. Whereas, the non-transitive part contains strategies that follow the conservation law; a typical example is the rock-paper-scissor game. 
The topological structure of real-world tasks often  resembles a spinning top if projected onto a 2D space  \citep{czarnecki2020real}, where the x-axis is the non-transitive dimension and y-axis is the transitive dimension. 
The non-transitive part can harm the effectiveness of auto-curriculum \citep{balduzzi2019open,czarnecki2020real}. For example, an auto-curriculum generated by self-play in zero-sum games \cite{gilpin1975limit,samothrakis2012coevolving} could make a learning agent  endlessly chase its own tail by creating the same tasks repetitively without breaking out. 
Things become even worse when the non-transitivity issue couples with the catastrophic-forgetting property of the model itself, as seen with deep neural networks \cite{kirkpatrick2017overcoming}. As a result,  agents may end up with acquiring mediocre solutions or getting trapped in limited cycles within the strategy space \cite{balduzzi2019open, balduzzi2018re}. 
Memorising a library of all possible policies can help prevent cycling (e.g., three strategies in the toy example of Rock-Paper-Scissors), but for many real-world tasks, the dimension of the non-transitive cycles can be huge, and as a result, building such a library itself becomes an endless task \citep{czarnecki2020real}.
%
%
%\begin{figure*}[t!]
%\vspace{-20pt}
%\includegraphics[width=0.9\linewidth]{diversity.pdf}
%\vspace{-10pt}
%\caption{ Generating diverse behaviour models for autonomous driving through auto-curriculum. }
%\label{fig:diversity}
%\vspace{-10pt}
%\end{figure*}

\vspace{-2pt}
\section{An Autonomous Driving Test Bed} %changed this
\label{sec:SMARTS}
Creating a diversity-aware auto-curricula for MARL, although still facing several open challenges, is a critical step for  deploying successful multi-agent learning systems in real-world domains. 
For validating effectiveness, we believe autonomous driving (AD) simulation environments provide an excellent test bed.

AD technologies \citep{badue2020self} enable a vehicle to sense its environment and move safely to a destination with little or no human intervention.
%An iconic event that spurred the development of self-driving technology was the DARPA Grand Challenge.
Since the first DARPA competition in 2004, where the best performing car completed only 7.3 miles of the 142-mile desert route, remarkable progress has been made. 
For example,  the commercial company Waymo  has driven more than $20$ million miles on public roads under the SAE level-4 setting \citep{sae2014automated,waymo2020safety}.
%The vision of deploying AD techniques for better safety and efficiency in transportation is gradually getting substantiated. 

In spite of such achievements, fully competent and natural interactions with other road users remain out-of-reach.
Rather than embracing inter-driver interaction, current mainstream level-4 AD solutions restricts it.
When encountering complex interactive scenarios, autonomous cars tend to slow down and wait for the situation to become more simple. They rarely cut in front of another car or force its way in at a merge, as human drivers routinely do.
In California in 2018, $86\%$ of crashes involving autonomous vehicles were attributable to the AD car's conservative behaviour \cite{stewart}, with $57\%$ rear endings and $29\%$ sideswipes by other vehicles on the AD car.
Trial AD cars in Arizona and California are often targets of complaints for blocking other cars \cite{Someofth6}, excessive hard braking \cite{WaymoRid1}, hesitant highway merging, and inflexible pick-up/drop-off locations \citep{WaymoB72,WaymoB80}.
While rarely illegal, the overly conservative driving style frustrates human drivers, and can even pose road hazards. This also restricts AD technologies from being applied on special-purpose vehicles, such as ambulances or police cars, where aggressive driving behaviour may be required.

A key reason for this limitation is that existing AD \textit{simulators} have limited capacity for modelling realistic interactions with diverse driving behaviours. 
Simulators are crucial for validation of the AI software controlling the autonomous vehicle (also called \emph{ego vehicle}).
For validating the ego vehicle's interactive behaviour with \emph{social vehicles} (i.e., other vehicles that share the same driving environment), we need diverse \emph{social agents} capable of realistic and competent interaction.  
Conventional AD simulators \cite{dosovitskiy2017carla, VTD:online, lgsvl:online} focus on modelling sensory inputs and control dynamics,  rather than interaction.
As a result, the behaviours of social vehicles end up being controlled by simple scripts or rule-based models (e.g., IDM  for longitudinal control \cite{treiber2000congested} or MOBIL for lateral control \cite{treiber2009modeling}) and the simulated interaction between ego and social vehicles falls short of the richness and diversity seen in the real world.
AD companies heavily use replay of historical data collected from real-world trials to validate ego vehicle behaviour \cite{InsideWa70}.
However, such a data-replay approach to simulation does not allow true interaction between vehicles because the social vehicles are not controlled by intelligent agents but merely stick to historical trajectories
\cite{DragoAng38, TheDisen76}.
In short, how to create a population of diverse intelligent social agents that can be adopted in simulation to provide traffic with realistic interaction is still an open question.

%Recently, data-driven approaches for training behaviour models capable of diverse and realistic interaction in simulators is receiving attention.
%Imitation learning is an obvious option \cite{ChauffeurNet}.
%Yet, a suitable data set is crucial.
%Publicly available AD datasets such as KITTI \cite{geiger2013vision}, Oxford RobotCar \cite{maddern20171}, and BDD100K \cite{yu2018bdd100k}
% focus on perception and provide little data regarding interaction.
%Although Argoverse \citep{chang2019argoverse}, nuScences \citep{nuscenes2019}, HDD dataset \citep{ramanishka2018toward}, and the 100-car study \citep{neale2005overview} have extended to prediction and thus could be used for imitation learning, they still suffer from incomplete labelling of the interacting entities and are limited to only the ego vehicle's perspective. 
%Bird's eye driving behaviour datasets, such as the NGSIM  \cite{alexiadis2004next} and highD \cite{krajewski2018highd} are limited because they focus on highways or traffic-light controlled intersections, with the INTERACTION dataset being a notable exception \citep{interactiondataset}. 
%%that includes a large number of roundabouts and unsignalized intersections.
%Moreover, all these data sets involve large scale human labelling or at a minimum require near perfect detection and tracking modules to generate the labels, hindering scaling and resource-constrained open research.
%In addition, they do not have the underlying physical models of the social vehicles, which could be crucial for generating realistic behaviours in highly dynamic situations.

We believe that a MARL approach powered by diversity-aware auto-curriculum can help solve the problem of generating high quality behaviour models that approach human-level sophistication.
Fundamentally, driving in a shared public space with other road users is a multi-agent problem wherein the behaviour of agents co-evolve.
Co-evolved diverse and competent behaviours
%\footnote{We argue that behavioural \emph{diversity} in the AD context should always be accompanied by a minimum \emph{quality} constraint in terms of safety and robustness. An accident should be avoided in all circumstances. This echoes to the Challenge III in Section \ref{sec:solutions}.} 
can allow AD simulation to encompass sophisticated interactions seen among human drivers and thus alleviate the conservativeness of existing AD solutions.  
Crucially, solving this problem for AD will also require the key research challenges identified in Section \ref{sec:solutions} to be addressed.

%
%\begin{figure}[t!]
%\vspace{-5pt}
%\includegraphics[width=0.99\linewidth]{smarts-arch-corl.pdf}
%\vspace{-5pt}
%\caption{SMARTS: this plot will be amended later}
%\label{fig:gdpp}
%\end{figure}
For such a plan to work, we need an appropriate simulator that supports MARL auto-curriculum for diversity, such as the SMARTS AD simulator \citep{zhou2020smarts}.
%provides a well-suited platform. 
Unlike other existing simulators, SMARTS is \emph{natively multi-agent}. Social agents use the same APIs as the ego agent to control vehicles, and thus  may use arbitrarily complex computational models, either rule-based or (MA)RL-driven.
The SMARTS \textit{social agent zoo} hosts a growing number of behaviour models to be used by simulated agents, regardless of their divergence in model architectures, observation/action spaces, or computational requirements.
The key component that makes such computations possible is the built-in ``bubble'' mechanism. This mechanism defines a spatiotemporal region so that  intensive  computing is  only activated inside the bubbles where fine-grained interaction is required, such as at unprotected left-turns, roundabouts, and highway double merges.
%In the meantime, agents, either ego or social, can learn from the interactive  trajectories collected within the bubbles, and adapt their behaviours accordingly. 
These features make SMARTS highly suitable for multi-agent auto-curriculum studies.

To be more concrete, consider the ULTRA \cite{ultra} benchmark suite built on top of SMARTS. It includes over 100,000 unprotected left-turn scenarios at different levels of complexity. 
These scenarios could seed an auto-curriculum that gradually increases the diversity and complexity of interaction by injecting trained behaviour models through available SMARTS mechanisms, allowing for not only more diverse agent behaviour, but also a curriculum composed of increasingly difficult scenarios.
Through this, we expect interaction behaviours reminiscent of the \emph{Pittsburgh Left} to emerge in a fashion that could potentially reach the level of sophistication of human drivers.
In turn, such emergent behavioural models can be used to support diverse interactions in AD simulations far beyond what has been possible through the rule-based models used so far.

\vspace{-2pt}
\section{Conclusion}
\label{sec:conclusion}
Despite remarkable success shown on video games, we envision that  developing a diversity-aware auto-curriculum framework is a critical step to ensure the success of  MARL technique on real-world problems.   
Specifically, we believe the pressing need for high-quality behaviour models in AD simulation is a great opportunity for the MARL community to make a unique contribution by (1) theoretically addressing the modelling challenges on behavioural diversity and (2) experimentally training generations of increasingly diverse and competent agents to provide interactions for real-world AD developments. %that is currently missing in AD simulation.
\clearpage
\bibliographystyle{ACM-Reference-Format} 
\balance
\bibliography{sample}

%%%%%%%%%%%%%%%%%%%%%%%%%%%%%%%%%%%%%%%%%%%%%%%%%%%%%%%%%%%%%%%%%%%%%%%%

\end{document}